# Integrating Testing and Operation-related Quantitative Evidences in Assurance Cases to Argue Safety of Data-Driven AI/ML Components

**Michael Kläs, Lisa Jöckel, Rasmus Adler, Jan Reich**

Fraunhofer Institute for Experimental Software Engineering IESE,
Fraunhofer-Platz 1, 67663 Kaiserslautern, Germany
{michael.klaes, lisa.joeckel, rasmus.adler, jan.reich}@iese.fraunhofer.de

**Abstract**

In the future, AI will increasingly find its way into systems that can potentially cause physical harm to humans. For such safety-critical systems, it must be demonstrated that their residual risk does not exceed what is acceptable. This includes, in particular, the AI components that are part of such systems' safety-related functions. Assurance cases are an intensively discussed option today for specifying a sound and comprehensive safety argument to demonstrate a system's safety. In previous work, it has been suggested to argue safety for AI components by structuring assurance cases based on two complementary risk acceptance criteria. One of these criteria is used to derive quantitative targets regarding the AI. The argumentation structures commonly proposed to show the achievement of such quantitative targets, however, focus on failure rates from statistical testing. Further important aspects are only considered in a qualitative manner – if at all. In contrast, this paper proposes a more holistic argumentation structure for having achieved the target, namely a structure that integrates test results with runtime aspects and the impact of scope compliance and test data quality in a quantitative manner. We elaborate different argumentation options, present the underlying mathematical considerations, and discuss resulting implications for their practical application. Using the proposed argumentation structure might not only increase the integrity of assurance cases but may also allow claims on quantitative targets that would not be justifiable otherwise.

## 1. Motivation

Components based on machine learning (ML) or artificial intelligence (AI) are increasingly necessary for many innovative, especially autonomous, systems, like self-driving vehicles in complex environments. They provide features that could not be realized (with competitive quality) using traditional software. It is almost unavoidable that these data-driven components (DDC) become safety-critical. Safety architectures can lower the criticality of a DDC but it is often hardly possible to get a DDC completely out of the safety-critical path. Currently, standards are missing for AI-enabled systems, but we need to find ways to assure that the AI-enabled system is sufficiently safe in its context.

Assurance refers to "grounds for justified confidence that a claim has been or will be achieved". Using assurance cases – as an established approach in safety engineering – also for this purpose is a heavily discussed option. An assurance case is defined as a "reasoned, auditable artefact created that supports the contention that its top-level claim (or set of claims) is satisfied, including systematic argumentation and its underlying evidence and explicit assumptions that support the claim(s)" [ISO/IEC/IEEE, 2019]. Assurance cases are a flexible means for dealing with standardization gaps and provide a structured way of arguing quality requirements of a (data-driven) software component. Using assurance cases for arguing safety if a DDC is in the safety-critical path is considered in current safety standardization (e.g., NWIP ISO/PAS 8800, VDE-AR-E 2842-61), research projects (e.g., KI Absicherung, EXAMAI), and communities (e.g., Safety-critical systems club).

In our previous work, we proposed a possible overall structuring of the argumentation of an assurance case based on complementary risk acceptance criteria [Kläs et al., 2021]. This means that there are two separate lines of assurance case argumentation, one following the "as low as reasonably practicable (ALARP)" risk acceptance criteria and the second one using a quantitative target, e.g., a sufficiently low probability for such AI outcomes that may affect safety.

Existing assurance case structures for arguing the quantitative targets mainly focus on evidences provided by statistical testing. Other important aspects contributing to achieving the quantitative target are not integrated in its argumentation but at most considered exclusively in qualitative way.

We see two problems with this: (a) On the one hand, ignoring such aspects as part of a quantitative argumentation may cause innovative systems not to be realized because the achievement of the quantitative safety targets of their DDCs cannot be argued considering testing evidences in isolation. Instead, it may be required to also consider evidences on runtime measures that are active at operation time to detect potentially wrong outcomes or scope compliance issues, such as safety supervisors [Feth at al., 2017] or uncertainty wrappers [Kläs and Sembach, 2019]. (b) On the other hand, ignoring relevant aspect as part of a quantitative argumentation can weaken the integrity of the assurance case and lead to overestimating system safety; for example, if the actual effect of scope incompliance or test data quality is ignored.

*Contribution:* To address the observed gap, we propose a more holistic quantitative argumentation that integrates both (a) evidences based on measures during testing and operation and (b) limitations due to scope incompliance and test data quality. Regarding data quality, we intentionally focus our example on label correctness, which is one of the most pressing issues in many practical applications today.

In this paper, we extend our previous work by providing (i) a flexible assurance case argument structure for quantitative safety targets on a data-driven component, (ii) the underlying mathematical model justifying the decomposition, and (iii) a discussion of the means and implication when applying the approach in practice.

The paper is structured as follows: Section 2 provides the background on assurance cases and our previous work on this topic. Section 3 presents related work on arguing quantitative targets. In Section 4, we incrementally develop an equation considering the different aspects for arguing the quantitative target. The paper concludes in Section 5.

## 2. Background and Previous Work

Existing safety standards do not define probabilistic target values for software because software fails in a systematic and not in a random way. For a concrete input, the software will deliver always the same output (if it is in the same state). This is also true for data-driven components (DDCs). In the following, we will thus first explain why probabilistic target values for DDCs make sense before we introduce our previous work on structuring the safety argument.

Safety standards define probabilistic target values for an overall safety function. If it is a continuous function like a collision avoidance function, then IEC 61508 refers to a target failure rate. If it is an on-demand function like manual safety shutdown, then it is a target failure probability. This makes it possible to derive probabilistic target values for parts of the system and prove that these targets are achieved. Even though target values are currently only used to address random faults, systematic faults also contribute to the achieved overall failure rate. It is reasonable to focus on fault avoidance, fault removal, and fault tolerance concerning systematic faults, but some systematic faults will remain in practice and could become active during operation. One approach to addressing this issue is to forecast the probability of activation and incorporate the evidences from fault forecasting into the argument for the target probability. Particularly for DDCs, this is a reasonable approach as (1) the likelihood of remaining systematic faults is high and (2) sophisticated techniques for estimating the probability of activating systematic faults exist.

Accordingly, in previous work [Kläs et al., 2021] we proposed deriving a probabilistic target value for a DDC failure and assuring by means of an assurance case that the target is achieved. This is illustrated in the right branch of the assurance case structure in Fig. 1. The target value could be taken from existing safety standards, which in turn derived it from the risk acceptance criterion Minimum Endogenous Mortality (MEM). Alternatively, one could derive it from the risk acceptance criterion Positive Risk Balance (PRB). In any case, we assume that such a target value exists for the overall safety function and that traditional safety analyses like fault tree analyses can be used to derive a target value for a DDC failure. The relevant development phases for assuring the achievement of this target are specification, testing, and operation. During construction and analyses, one can avoid and remove faults but not generate evidences concerning the activation of residual faults. Construction and analyses are thus necessary to achieve the target but are not used for arguing the achievement based on evidences.

At least theoretically, construction and analyses could avoid and remove so many faults that a much better value is achieved than the target value. In practice, this could occur if a relatively high failure probability is acceptable. Imagine, for instance, a safety warning to avoid operator failures that impose a very low risk. The risk is so low that it has been traditionally accepted and not addressed with a safety warning. One could argue that a safety warning with a high probability of omission is better than no warning. Consequently, the probabilistic target value would be significantly less demanding than the values defined in IEC 61508 for safety functions. In such cases, we want to argue that faults are avoided and removed as far as reasonably practicable and not only that target values are achieved. This argument is reflected in the complementary left branch in Fig. 1.

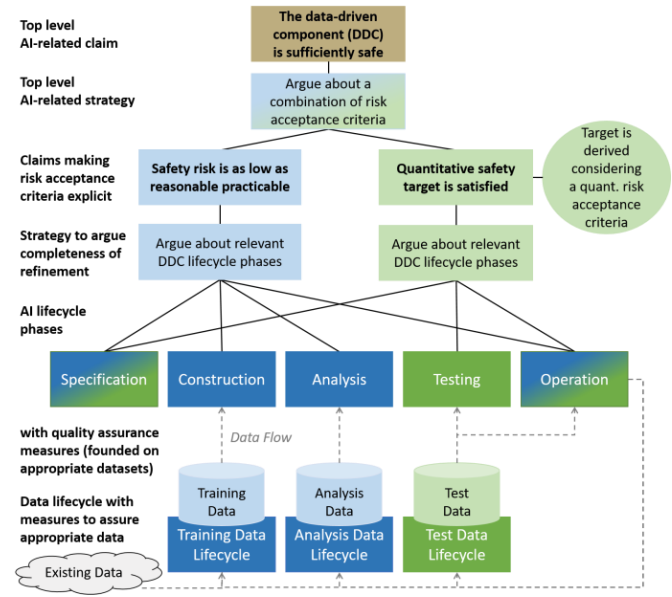

Fig. 1: Assurance case with complementary risk acceptance criteria [Kläs et al., 2021]

In previous work, we proposed this structure but did not go into details of the two branches. In this paper, we will refine the right branch and focus on (a) evidences based on measures during testing and operation as well as (b) limitations due to scope incompliance and test data quality.

## 3. Related Work

This section outlines how existing work on assurance case structures for DDCs argue quantitative targets, i.e., whether and how they consider evidences from statistical testing, runtime uncertainty estimation, or data quality assessment.

Most argumentation structures consider evidences from testing during design time as understood by Jöckel et al. [2021] (e.g. [Mock et al., 2021], [Hawkins et al., 2021]). Runtime mechanisms contributing to quantitative target achievement are so far being considered in a qualitative manner, i.e., proposing to consider uncertainties during runtime, or implicitly quantitative like require a sufficient monitoring of erroneous inputs, outputs, and assumptions including the uncertainty associated with the outcomes of the model [Hawkins et al., 2021].

Mock et al. [2021] consider deep neural network measures that are applied either during design time or runtime, including uncertainty estimation, and testing measures. How the evidences provided by these measures are used in a safety argumentation is not detailed. Cheng et al. [2019] introduced the nn-dependability-kit for arguing the quality of neural networks. It provides algorithms to produce evidences for indicating sufficient elimination of uncertainties in the product lifecycle, a reasoning engine to ensure that the generalization does not lead to undesired behaviors, and runtime monitoring for reasoning on whether a decision of a neural network in operation is supported by prior similarities in the training data. Although the provided GSN argument relates the techniques and metrics to a quantitative top claim, it is neither argued why the solutions in their entirety imply achievement of the quantitative target nor is a concrete mathematical basis given for combining design-time and runtime techniques.

Some works on assurance cases for AI are based on the lifecycle stages and associated methods proposed by Ashmore et al. [2019], e.g., the work of Picardi et al. [2020] and Hawkins et al. [2021]. They address testing as part of the model verification stage and state that providing a suitable measure of confidence in ML model output, i.e., the level of uncertainty in the model outcome, is an open challenge of the model deployment stage, which also extends to the model learning and model verification stages. However, testing and runtime uncertainty estimation are so far seen as separate activities in the quality assurance of a DDC. How they contribute to achieving the overall quantitative safety target in combination is not stated.

Following an argumentation structure in line with ISO 26262 and common strategies for fault elimination and fault tolerance, Rudolph et al. [2018] propose splitting the argument into techniques for redundancy patterns and online monitoring for the realization of DDC software and design-time techniques for eliminating DDC hazards. The proposed evidence types represent a set of practices that can be used, but no (mathematical) argument is given for why they are sufficient to justify safety.

Burton et al. [2019] introduce confidence arguments that require a contract performance, which might be understood as a quantitative target, for which a measurement target needs to be exceeded by a provided evidence. They describe rather the pattern of confidence arguments, which might be compatible with our proposed argumentation structuring.

Regarding the consideration of data quality as part of the argumentation of a quantitative target, the quality or the appropriateness of the data is commonly described as a challenge (e.g. [Gabreau et al., 2021], [Delseny et al., 2021]) but is kept on a broader level in reviewed assurance frameworks. Schwalbe and Schels [2020] propose input space coverage for the development data and additional coverage of model behavior and available experience for test data.

Wozniak et al. [2020] integrate goals that support the data appropriateness claim concerning the data satisfying the ML safety requirements, proper labeling, domain coverage, amount, and disjointness. The AMLAS report by Hawkins et al. [2021] includes evidences in the assurance argument regarding the sufficiency of the data meeting the ML data requirements, which consider the relevance, completeness, accuracy, and balance of the data. Discrepancies between the data and the ML data requirements need to be justified and documented.

In summary, the reviewed proposals assume either that an appropriate dataset is given or quality deficits should be documented, like in the AMLAS report. However, no possibility of estimating the quality deficits and considering their impact in the performance estimation of the DDC has been presented yet to the best of our knowledge.

## 4. Framework to Argue Quantitative Target

In this section, we will first introduce some basic terminology und discuss the meaning of a quantitative safety target for a data-driven component (DDC). Next, we will incrementally develop an equation considering the different factors that are relevant when determining whether a given quantitative safety target for a DDC is achieved and illustrate their application and impact on an example. Finally, we will make a proposal how to argue the achievement of a quantitative target based on the derived equation using an assurance case for which we provide a generic structure in Fig. 2. The proposed structure extends and breaks down the quantitative argument on the right side of Fig. 1, which considers additionally ALARP on the left side as a complementary risk acceptance criterion [Kläs et al., 2021].

### (A) Quantitative Safety Targets for DDC

A DDC is a component that makes use of a data-driven model (DDM), i.e., a model derived from data using ML or another data-driven AI approach. If a DDC becomes part of safety-related functions, the quantitative safety target defined for this function based on the chosen quantitative risk acceptance criterion (e.g., Minimum Endogenous Mortality or Positive Risk Balance) would delegate some of the safety responsibility to the DDC. This means there is also a quantitative safety target for the DDC. Based on the chosen architecture and the information sources considered, this safety target may differ. Regardless of its concrete definition, it will be quantitative, probabilistic, and commonly request that *DDC-caused safety violations* are sufficiently unlikely (considering a specified level of confidence). This means, there is a probability target $p_{target}$ that must not be exceeded on a given level of confidence $CL$. We like to remark that not every DDC-caused safety violation will also cause a hazard on the system boundary, but the probability of a hazard is commonly further reduced by safety measures on the system level, e.g., by fusing information sources.

For illustrative purposes, we consider the following simplified example: The considered safety function is responsible for assuring that a vehicle stops at intersections where there is a stop sign, and the DDC we consider is a traffic sign recognition component that is applied to decide based on a given image whether there is a stop sign or not. A *DDC-caused safety violation* will occur if the DDC provides some information that might result in crossing the intersection without stopping although there is a stop sign. This occurs if the DDC provides information that it is sure that there is no stop sign, which means it produces a *safety-related failure* (SRF) by providing the false negative information that there is no stop sign and additionally does not mark this information as potentially wrong. Thus, the quantitative safety target needs to be refined for the DDC into something like 'the probability of a SRF of the DDM provided by the DDC without a warning is less than $p_{target} = 0.002$'. Note that this means that not every wrong outcome of the DDM automatically represents a DDC-caused safety violation. Such a violation only occurs if the wrong outcome is a SRF, i.e., a false negative in this case, and if the DDC does not inform the system about the potentially wrong outcome.

To increase trust in the provided evidence on the achievement of this target, we may further request a confidence level of $CL = 0.9999$ for statistics that are applied to indicate that the probability of a DDC-caused safety violation $p_{\overline{safe}}$ is less or equal $p_{target}$.

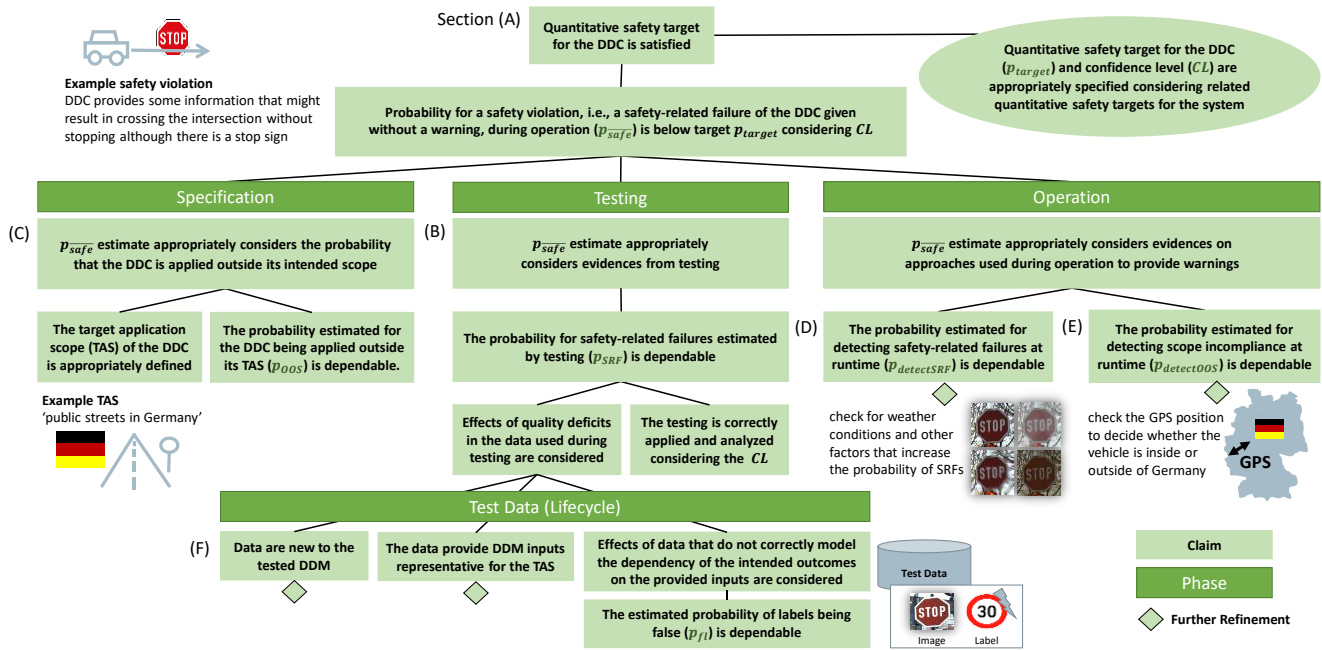

Fig. 2 Proposal on an assurance case structure to integrate relevant factors to argue a quantitative safety target

**(B) Statistical Testing**

In order to determine $p_{\overline{safe}}$ and show that $p_{\overline{safe}} \leq p_{target}$, the most obvious and common approach is testing. If we can assure, e.g., by organizational means that the DDC is only applied in its intended way and there are no changes in the context, we can assume that the probability of the model being applied outside its *target application scope* (TAS), also named ODD, is zero, i.e., the probability of out-of-scope applications $p_{OOS} \stackrel{\text{def}}{=} p(\overline{TAS}) = 0$, where $\bar{A}$ compliments $A$.

$$p_{\overline{safe}} = p(SRF) = p(SRF|TAS)\,p(TAS) + p(SRF|\overline{TAS})\,p(\overline{TAS}) = p(SRF|TAS) \quad (1)$$

In this case, $p(SRF|TAS)$ is the conditional probability that the outcome of the DDM leads to a safety-related failure *SRF* when applied for inputs from its TAS.

Assuming that the TAS in our simplified example is 'public streets in Germany', $p(SRF|TAS)$ is the probability that the DDM does not recognize the stop sign when applied on images with a stop sign taken on public streets in Germany. This is what the onion shell model of uncertainty considers as *model-fit-related uncertainty* [Kläs and Vollmer, 2018].

The probability $p(SRF|TAS)$ can be estimated in this case from testing results applying statistical testing on a test dataset with perfect quality that contains a representative sample of images taken from public streets in Germany with stop signs. Note that since in this case, we test for safety-related failures, no images without a stop sign are required. Specifically, we determine $p_{testSRF}$ as an upper bound on the acceptable safety-related failure rate on the test dataset based on a binomial proportion confidence interval, so that $p(SRF|TAS) \leq p_{testSRF}$ with probability $CL$, we denote as $p(SRF|TAS) \leq_{CL} p_{testSRF}$. Thus, we achieved the safety target for DDC with confidence $CL$ if $p_{testSRF} \leq p_{target}$.

In our example, we thus need to show $p_{testSRF} \leq 0.002$ considering $CL = 0.9999$. More specifically, assuming a test dataset with 100.000 images with stop signs, we need to check that the number of false negatives observed during testing is less than 149.

**(C) Target Application Scope Compliance**

In the previous section, we assumed that we could assure that the DDC is never applied outside its TAS. However, in most real-world applications, this assumption is not justifiable. In our example, we might have situations where the vehicle crosses the border to France at a certain point in time, or stop signs in Germany might get an updated design. In addition, the traffic scenes in which stop signs need to be detected may change over the years due to new types of buildings, road users, etc. This is what the onion shell model of uncertainty considers as *scope-compliance-related uncertainty* [Kläs and Vollmer, 2018]. Under the assumption that $p_{OOS} > 0$, we need to include the probability of out-of-scope applications, reconsidering Eq. 1

$$p_{\overline{safe}} \leq p(SRF|TAS)(1 - p_{OOS}) + p(SRF|\overline{TAS})\,p_{OOS} \quad (2)$$

Because there are infinitely many inputs outside the TAS for which we did not and cannot reasonably test the DDM in a representative way, we need to make the worst-case approximation that $p(SRF|\overline{TAS}) = 1$ and obtain from Eq. 2

$$p_{\overline{safe}} \leq p(SRF|TAS)(1 - p_{OOS}) + p_{OOS} \quad (3)$$

Unlike $p(SRF|\overline{TAS})$, $p_{OOS}$ can be estimated. Depending on the planned application, this can be done, for example, by using previous data collected on the usage of the system in which the DDM is planned to be applied and on the experience of domain experts. Usually, there is no possibility to derive $p_{OOS}$ solely statistically. The expert estimates taken into accounts should be sufficiently pessimistic considering the stated confidence level $CL$. Details on what to consider when using expert estimates and how to deal with related bias and uncertainty can be found in the literature (e.g., [O'Hagan et al., 2006]). We obtain and simplify from Eq. 3

$$p_{\overline{safe}} \leq_{CL} p_{testSRF}(1 - p_{OOS}) + p_{OOS} = (1 - p_{\overline{testSRF}})\,p_{\overline{OOS}} + (1 - p_{\overline{OOS}}) = 1 - p_{\overline{testSRF}}\,p_{\overline{OOS}} \quad (4)$$

In our example, we derive a monotonic increasing function $p_{OOS}(t)$ over time t that gives us an estimate of $p_{OOS}$ considering available fleet data on past vehicle locations and recorded images as well as the update rate observed for traffic signs in the past. Considering a specific t, we can obtain $p_{OOS}(t) = 0.0005$. In this case, we need to show during testing that $p\_testSRF \leq (0.002 - 0.0005)/(1 - 0.0005) \approx 0.0015$ compared to 0.002 in the case (C).

**(D) Runtime Detection of Safety-related Failures**

In recent years, an increasing number of papers have proposed considering approaches such as uncertainty estimation [Kläs et al., 2021a] and safety supervisor architectures [Feth et al., 2017] to help safeguard DDMs during operation. At the core of these works is the idea to detect DDM outcomes at runtime that potentially cause safety-related failures and report them to the system level where they are handled. Measures on the system level include, for example, switching to a second channel with a less effective but safety-proven algorithm or taking some safety actions such as reducing velocity, requesting takeover by the human driver, or a minimal risk maneuver [Feth et al., 2017].

In consequence, DDM outcomes detected as potentially causing safety-related failures and marked by the DDC as such will reduce the probability of a safety violation, i.e., reduce $p_{\overline{safe}}$. Assuming that we can show that the approach detects a safety-related failure when applied in the TAS with the probability of $p(detect(SRF)|SRF \wedge TAS)$, we can extend Eq. 3:

$$p_{\overline{safe}} \leq p(SRF|TAS)(1 - p_{OOS}) + p_{OOS} - p(SRF|TAS)\,p(detect(SRF)|SRF \wedge TAS) \quad (5)$$

With $1 - p_{OOS} \geq p(detect(SRF)|SRF \wedge TAS)$ as a preposition, which commonly holds in most practical applications because the probability of out-of-scope applications is usually rather small, we can replace $p(SRF|TAS)$ with $p_{testSRF}$ in Eq. 5, keeping the inequality with confidence level $CL$:

$$p_{\overline{safe}} \leq_{CL} p_{testSRF}(1 - p_{OOS}) + p_{OOS} - p_{testSRF}\, p(detect(SRF)|SRF \wedge TAS) \quad (6)$$

Similar to the way in which the probability of safety-related failure can be estimated by testing on a representative sample from TAS, we can also determine the probability of detecting failures at operation. To estimate $p(detect(SRF)|SRF \wedge TAS)$, we determine the lower bound considering the ratio of safety-related failures that are detected on a representative sample with the given level of confidence to obtain $p_{detectSRF}$ with $p(detect(SRF)|SRF \wedge TAS) \leq_{CL} p_{detectSRF}$, resulting in

$$p_{\overline{safe}} \leq_{CL} p_{testSRF}(1 - p_{OOS} - p_{detectSRF}) + p_{OOS} \quad (7)$$

This means in our example with $p(SRF|TAS) \approx 0.002$ and an overall sample size of 100,000 images that we have approximately 200 samples that will result in safety-related failures. Assuming that on this sample, the approach detects 85 of the safety-related failures, we would get $p_{detectSRF} \approx 0.30$ on $CL = 0.9999$. This would relax the requirement on testing-based evidence to $p_{testSRF} \leq (0.002 - 0.0005)/(1 - 0.0005 - 0.30) \approx 0.0021$ versus 0.0015 in case (D).

**(E) Runtime Out-of-Scope Detection**

It is commonly hard or even impossible to assure that a DDC is only applied within its TAS. Therefore, we take measures to detect out-of-scope applications. An example of such an approach is SafeML [Aslansefat et al., 2020], but even simpler approaches such as checking sensor data for boundary conditions can help to detect if the DDC is applied outside its TAS.

In the considered example, we can check the GPS position to decide whether the vehicle is inside or outside of Germany, or determine and evaluate the "similarity" of the recent inputs at operation to the inputs considered during testing.

Assuming that we can detect out-of-scope applications with the probability of $p_{detectOOS} \stackrel{\text{def}}{=} p(detect(\overline{TAS})|\overline{TAS})$, we can extend Eq. 5:

$$p_{\overline{safe}} \leq p(SRF|TAS)(1 - p_{OOS}) + p_{OOS} - p(SRF|TAS)\, p(detect(SRF)|SRF \wedge TAS) - p_{OOS}\, p_{detectOOS} \quad (8)$$

To the best of our knowledge, there is no reasonable way to justify the performance of advanced out-of-scope detection approaches since from a theoretical point of view, they perform a one-class-classification task [Moya and Hush, 1996]. This means we do not have a representative dataset with data points labeled as being inside/outside the TAS but only one for data points from inside the TAS. Therefore, we can determine their false alarm rate and compare their relative performance on datasets, but will get no absolute numbers on their out-of-scope detection performance [Kläs and Jöckel, 2020]. We thus clearly recommend their use for DDCs but currently consider it questionable to make use of them in a quantitative argument. We rather consider them as the last line of defense that catches at least some of the issues whose consideration has been missed. On the other hand, there are out-of-scope detection means with quantifiable effects. In our example, using the GPS location can clearly reduce the probability that the DDC is used when the vehicle is outside of Germany. The effect of such detection measures can be estimated based on available data and expert opinion at design time and considered in a quantitative argument. Making the same assumptions as before, we obtain from Eq. 8

$$p_{\overline{safe}} \leq_{CL} p_{testSRF}(1 - p_{OOS} - p_{detectSRF}) + p_{OOS}(1 - p_{detectOOS}) \quad (9)$$

Assuming that based on our TAS analysis, the probability that the DDC is used in a vehicle that is currently outside of Germany contributes to 50% of the out-of-scope cases and we can detect this (conservatively) in 99% of the cases. Considering no further aspects regarding out-of-scope detection, we get $p_{detectOOS} = 0.495$. This would relax the requirement on testing-based evidence to $p_{testSRF} \leq (0.002 - 0.0005(1 - 0.495))/(1 - 0.0005 - 0.30) \approx 0.0025$.

**(F) Consideration of Data Quality**

When we estimate the failure probabilities of a DDM through statistical testing and detection probabilities for runtime monitoring, the validity of our results depends on assumptions on the quality of the data used to determine these results. Specifically, (a) the used dataset *ds* has to be unseen, e.g., it was not used during DDM development to train or optimize the model; (b) it has to provide inputs that are representative of the target application scope *TAS*, i.e., $P(input_{ds}) \sim P(input_{TAS})$; and (c) it has to correctly model the relationship between inputs and intended outcomes $P(outcome_{ds}|input_{ds}) \sim P(outcome_{TAS}|input_{TAS})$.

In reality, the validity of these assumptions is threatened by various quality issues, e.g., an auto correlation, a biased data collection, or labeling errors. Thus, these assumptions need to be modeled as claims and justified by evidences.

An obvious way would be to address the quality issues using an ALARP argumentation. Yet, considering just an ALARP argumentation makes the assumption that doing as much as reasonably practical will reduce existing issues to a magnitude where their impact can be neglected in the argumentation of the quantitative safety target. However, experience from many data science projects and in-depth discus-sions with practitioners indicate that for most real-world settings, such a level of data quality would require investments far beyond what can be considered reasonably practical.

In consequence, limitations regarding data quality as stated in data-quality-related claims need to be captured in a quantitative manner and integrated into the argumentation of the quantitative safety target.

Quantifying and providing operationalization for all kinds of quality issues affecting the three data quality claims summarized in Figure 2, however, exceeds the scope of this paper. Thus, we illustrate the basic idea and applicability of our proposal on one of the most pressing quality issues in many practical applications, namely dealing with datasets containing a significant number of incorrectly labeled data.

We define the probability of a labeling fault as $p_{lf} := p(l_{ds} \neq l \mid TAS)$, i.e., the probability that the label $l_{ds}$ of a random data point in the dataset $ds$ does not match the label $l$ that one would expect considering the intended purpose of the DDM. $p_{lf}$ can usually be empirically approximated with reasonable effort using statistics, e.g., [Eagan et al., 2020] [Kulesza et al., 2014], considering annotator agreement between multiple labelers or repeated measurements.

Since cases that are more difficult to label correctly are commonly also cases that have a higher chance of a wrong DDC outcome, it does not appear reasonable to assume that $p(SRF \mid l_{ds} = l \wedge TAS) = p(SRF \mid l_{ds} \neq l \wedge TAS)$.

Yet, we can decompose the probability of an actual SRF within the TAS using the law of total probability and provide an upper bound given that $SRF_{ds}$ comprises outcomes rated as SRF based on the dataset $ds$:

$$p(SRF|TAS) = p(SRF \wedge SRF_{ds} |TAS) +$$
$$p(SRF \wedge \overline{SRF_{ds}} |TAS) \leq p(SRF \wedge SRF_{ds} |TAS) + p_{lf} \quad (10)$$

under the safety-relevant assumption that, in case of a labeling fault, an outcome of the DDM is not recognized as a safety-related failure on the dataset (due to the labeling error), but is actually a safety-related failure.

Bayes' theorem further gives us:

$$p(SRF \wedge SRF_{ds} |TAS) =$$
$$p(SRF \mid SRF_{ds} \wedge TAS) * p(SRF_{ds}|TAS) \quad (11)$$

where $p(SRF \mid SRF_{ds} \wedge TAS)$ is the probability that outcomes considered as safety-related failures based on the dataset $ds$ are actually safety-related failures. As we assume that this is the case for all labeling faults, we have $p(SRF \mid SRF_{ds} \wedge TAS) = 1$.

Hence, we can estimate the probability of a safety-related failure based on test data and the labeling faults:

$$p(SRF|TAS) \leq p(SRF_{ds}|TAS) + p_{lf}. \quad (12)$$

As $p_{testSRF_{ds}} \leq_{CL} p(SRF_{ds}|TAS)$ can be obtained using statistical testing on $ds$, we can use (12) to integrate the labeling fault into the equations in the previous section. For case (B), we get $p_{\overline{safe}} \leq_{CL} p_{testSRF_{ds}} + p_{lf}$, leading to an acceptable number lower than 64 (instead of 149) false negatives in the example with an assumed $p_{lf} = 0.001$. For the example in (C), we would require a safety-related failure probability determined by testing that is less than 0.0005 instead of 0.0015, similarly for (D), with 0.0011 instead of 0.0021, and for (E), with 0.0015 instead of 0.0025.

**Argument Structure for Quantitative Targets**

In the previous subsection, we derived and illustrated equations that allow combining various pieces of information to show that a quantitative safety target defined as a maximum probability for DDC-caused safety violations is achieved. This subsection makes a proposal how to arrange the provided pieces in an argument structure using assurance cases.

The objective of the paper is not to a provide a formally specified assurance case, but rather to illustrate the structure of argumentation we propose. Thus, we did not apply a specific assurance case notation such as GSN [SCSC, 2018] but focus in our overview given in Fig. 2 on the breakdown of the claims and their mapping to the DDC lifecycle. We also intentionally neglected additional argument elements in Fig. 2 but explain the applied strategies in subsections (A) to (F).

Our top-level claim is taken from the assurance case given in the background section (Fig. 1) considering the branch arguing the quantitative target. In a first refinement, the top-level claim is split into (A) the claim that we appropriately derived the quantitative target for the DDC from the quantitative target for the system and the claim that the DDC target is achieved. As we have shown in Eq. 7, providing a dependable estimate on the probability of safety violations depends on several factors, we provide corresponding sub claims. These sub claims address the dependability of the estimates on (B) the probability of safety-related failure as determined during testing, (C) the probability of scope compliance derived during specification, and the effectiveness of detection mechanisms applied at operation, where we distinguish between (D) detecting safety-related failures for applications in the TAS and (E) detecting out-of-scope applications.

As the dependability of the before mentioned estimates depends on the data they use, data quality is considered in (F). There, we illustrate for a common type of data quality issue how to include observed limitations in the calculation using Eq. 12 and the corresponding argument structure.

## 5. Conclusion

In this paper, we make a proposal how to argue by means of a mathematical foundation that a DDC has achieved a given quantitative safety target. The proposal considers and integrates quantitative evidences from statistical testing, runtime monitoring, data quality assessment, and anticipated scope compliance.

This integration closes a gap in existing argument structures, up to our knowledge, since existing approaches for arguing the achievement of a quantitative safety target consider only single aspects in isolation but do not mathematically integrate them. As illustrated in the provided example, we see the chance to strengthen the safety argument by means of this integration and work with better assumptions on data quality. Particularly, the integration of the aspect "data quality" strengthens the safety argument for the quantitative target, as data quality is a major issue in practice. Our approach and its mathematical foundation cannot only be used in context of goal-based safety standards and related assurance cases but also for developing safety metrics for rule-based safety standards.

In a next step, we plan to instantiate the structure on several real-world applications to evaluate its feasibility and usefulness. Moreover, we plan to extend the presented equations and claims to address further data quality aspects including common issues like biased and incomplete datasets.

## Acknowledgments

Parts of this work have been funded by the Observatory for Artificial Intelligence in Work and Society (KIO) of the Denkfabrik Digitale Arbeitsgesellschaft in the project "KI Testing & Auditing", by the German Federal Ministry of Education and Research (BMBF) in the project "DAITA", and by the projects "AIControl" and "LOPAAS" as part of the internal funding programs of Fraunhofer.

## References

[Ashmore et al., 2019] R. Ashmore, R. Calinescu, and C. Paterson. Assuring the Machine Learning Lifecycle: Desiderata, Methods, and Challenges, *ACM Computing Surveys,* 2019.

[Aslansefat et al., 2020] K. Aslansefat, I. Sorokos, D. Whiting, R. Tavakoli Kolagari, Y. Papadopoulos. SafeML: Safety Monitoring of Machine Learning Classifiers Through Statistical Difference Measures. In: *Proc. of IMBSA 2020*, 2020.

[Burton et al., 2019] S. Burton, L. Gauerhof, B. B. Sethy, I. Habli, and R. Hawkins. Confidence Arguments for Evidence of Performance in Machine Learning for Highly Automated Driving Functions. In Proc. of *SAFECOMP 2019*, 2019.

[Cheng et al., 2019] C.-H. Cheng, C.-H. Huang, and G. Nührenberg. nn-dependability-kit: Engineering Neural Networks for Safety-Critical Systems. arXiv:1811.06746v2, 2019.

[Delseny et al., 2021] H. Delseny, C. Gabreau, A. Gauffriau, B. Beaudouin, L. Ponsolle, et al. White Paper Machine Learning in Certified Systems. arXiv:2103.10529v1, 2021.

[Eagan et al., 2020] B. Eagan, J. Brohinsky, J. Wang, and D. W. Shaffer. Testing the reliability of inter-rater reliability. In *Proc. of Learning Analytics & Knowledge (454-461), 2020*.

[Feth et al., 2017] P. Feth, D. Schneider, R. Adler. A conceptual safety supervisor definition and evaluation framework for autonomous systems. In *Proc. of SAFECOMP 2017*, 2017.

[Gabreau et al., 2021] C. Gabreau, B. Pesquet-Popescu, F. Kaakai, and B. Lefevre. Artificial Intelligence for Future Skies: On-going Standardization Activities to Build the Next Certification/Approval Framework for Airborne and Ground Aeronautic Products. In *Proc. of AISafety 2021, 2021.*

[Hawkins et al., 2021] R. Hawkins, C. Paterson, C. Picardi, Y. Jia, R. Calinescu, and I. Habli. *Guidance on the Assurance of Machine Learning in Autonomous Systems (AMLAS), 2021.*

[ISO/IEC/IEEE, 2019] ISO/IEC/IEEE 15026-1:2019 – Systems and software engineering - Systems and software assurance - Part 1: Concepts and vocabulary, 2019.

[Jöckel et al., 2021] L. Jöckel, T. Bauer, M. Kläs, M. P. Hauer, and J. Groß. Towards a Common Testing Terminology for Software Engineering and Data Science Experts. In *Proc. of Profes 2021*, 2021.

[Kläs and Jöckel, 2020] M. Kläs und L. Jöckel. A Framework for Building Uncertainty Wrappers for AI/ML-based Data-Driven Components. In Proc. of *SAFECOMP 2020*.

[Kläs and Sembach, 2019] M. Kläs, and L. Sembach. Uncertainty Wrappers for Data-Driven Models. In *Proc. of SAFECOMP 2019*. Springer, pp. 358–364, 2019.

[Kläs and Vollmer, 2018] M. Kläs, and A.M. Vollmer. Uncertainty in Machine Learning Applications: A Practice-Driven Classification of Uncertainty. In Proc. of *SAFECOMP 2018*.

[Kläs et al., 2021] M. Kläs, R. Adler, L. Jöckel, J. Groß, and J. Reich. Using Complementary Risk Acceptance Criteria to Structure Assurance Cases for Safety-Critical AI Components. In *Proc. of AISafety 2021*, 2021.

[Kläs et al., 2021a] M. Kläs, R. Adler, I. Sorokos, L. Joeckel, J. Reich. Handling Uncertainties of Data-Driven Models in Compliance with Safety Constraints for Autonomous Behavior. In *Proc. of European Dependable Computing Conference (EDCC) 2021*, 2021.

[Kulesza et al., 2014] T. Kulesza, S. Amershi, R. Caruana, D. Fisher, and D. Charles. Structured labeling for facilitating concept evolution in machine learning. In *Proc. of SIGCHI Conference on Human Factors in Computing Systems (pp. 3075-3084),* 2014.

[Mock et al., 2021] M. Mock, S. Scholz, F. Blank, F. Hüger, A. Rohatschek, L. Schwarz, and T. Stauner. An Integrated Approach to a Safety Argumentation for AI-Based Perception Functions in Automated Driving. In *Proc. of SAFECOMP 2021,* 2021.

[Moya and Hush, 1996] M. Moya, D. R. Hush. Network constraints and multi-objective optimization for one-class classification. *Neural Networks* 9(3), 463-474, 1996.

[O'Hagan et al., 2006] A. O'Hagan, A., et al. *Uncertain Judgements: Eliciting Expert Probabilities*. John Wiley, 2006.

[Picardi et al., 2020] C. Picardi, C. Paterson, R. Hawkins, R. Calinescu, and I. Habli. Assurance Argument Patterns and Processes for Machine Learning in Safety-Related Systems. In *Proc. of SafeAI 2020*, pp. 23-30, 2020.

[Rudolph et al., 2018] A. Rudolph, S. Voget, and J. Mottok. A consistent safety case argumentation for artificial intelligence in safety related automotive systems. In *Proc. of ERTS 2018*.

[Schwalbe and Schels, 2020] G. Schwalbe, and M. Schels. A Survey on Methods for the Safety Assurance of Machine Learning Based Systems. In: *Proc. of European Congress on Embedded Real Time Software and Systems*, 2020.

[SCSC, 2018] Safety-Critical Systems Club. *GSN Community Standard Version 2 Draft 1*, 2018.

[Wozniak et al., 2020] E. Wozniak, C. Cârlan, E. Acar-Celik, and H. Putzer. A Safety Case Pattern for Systems with Machine Learning Components. In *Proc. of SAFECOMP 2020*. Springer, pp. 370-382, 2020.